\documentclass[letterpaper,twocolumn,10pt]{article}
\usepackage{usenix-2020-09}

\usepackage{tikz}
\usepackage{amsmath}

\usepackage{tabu}                      
\usepackage{booktabs}                  
\usepackage{tabularx}
\usepackage{graphicx}
\usepackage{subcaption}
\usepackage{lipsum}                    
\usepackage{mwe}                       
\usepackage{algorithm}
\usepackage{algpseudocode}
\usepackage{authblk}

\usepackage{filecontents}

\begin{document}

\date{}

\title{\Large \bf GS-Cache: A GS-Cache Inference Framework for Large-scale Gaussian Splatting Models}


\author[1*]{Miao Tao}
\author[1*]{Yuanzhen Zhou}
\author[1*]{Haoran Xu}
\author[1]{Zeyu He}
\author[1]{Zhenyu Yang}
\author[1]{Yuchang Zhang}
\author[1]{Zhongling Su}
\author[1]{Linning Xu}
\author[1]{Zhenxiang Ma}
\author[1]{Rong Fu}
\author[1]{Hengjie Li}
\author[1]{Xingcheng Zhang}
\author[2]{Jidong Zhai}

\affil[1]{Shanghai Artificial Intelligence Laboratory, Shanghai, China}
\affil[2]{Tsinghua University, Beijing, China}

\maketitle

\begin{abstract}
Rendering large-scale 3D Gaussian Splatting(3DGS) model faces significant challenges in achieving real-time, high-fidelity performance on consumer-grade devices. Fully realizing the potential of 3DGS in applications such as virtual reality (VR) requires addressing critical system-level challenges to support real-time, immersive experiences. We propose GS-Cache, an end-to-end framework that seamlessly integrates 3DGS’s advanced representation with a highly optimized rendering system. GS-Cache introduces a cache-centric pipeline to eliminate redundant computations, an efficiency-aware scheduler for elastic multi-GPU rendering, and optimized CUDA kernels to overcome computational bottlenecks. This synergy between 3DGS and system design enables GS-Cache to achieve up to 5.35x performance improvement, 35\% latency reduction, and 42\% lower GPU memory usage, supporting 2K binocular rendering at over 120 FPS with high visual quality. By bridging the gap between 3DGS’s representation power and the demands of VR systems, GS-Cache establishes a scalable and efficient framework for real-time neural rendering in immersive environments.
\end{abstract}

\renewcommand{\thefootnote}{}
\footnotetext{* denotes equal contribution.}

\section{Introduction}

Real-time rendering of high-quality, large-scale 3D scenes is a resource-intensive task. The rendering of 3D scenes plays a crucial role in numerous fields, including virtual reality (VR), augmented reality (AR), and the metaverse. As novel methods for 3D reconstruction and rendering, neural radiance fields (NeRF)~\cite{Mildenhall2020NeRF} and 3DGS~\cite{Kerbl20233DGS} are capable of rendering incredibly photo-realistic and detailed images, supporting a variety of applications with high visual perception requirements. As the demands for finer scene quality and larger scene scale increase, new methods such as structured Gaussian derivation methods~\cite{lu2024scaffold, ren2024octree} continue to emerge. 3DGS methods have enabled the reconstruction of large-scale scenes, achieving the city-level 3DGS model training~\cite{kerbl2024hierarchical, liu2025citygaussian, lin2024vastgaussian}. However, the computation and memory intensity of real-time rendering in large-scale scenes increases significantly with the expansion of scene scale, and general solutions usually involve a scaling up~\cite{zhao2024grendal} of computing resources. The structured Gaussian derivation method achieves high-quality reconstruction and reduces the difficulty of real-time rendering in large-scale scenes to a certain extent from the model structure, which makes it one of the most promising methods for achieving photo-realistic rendering of large-scale scenes for VR. Because the demands for quality and performance in VR rendering are significantly higher than in other applications.

\begin{figure}
    \centering
    \includegraphics[width=1\linewidth]{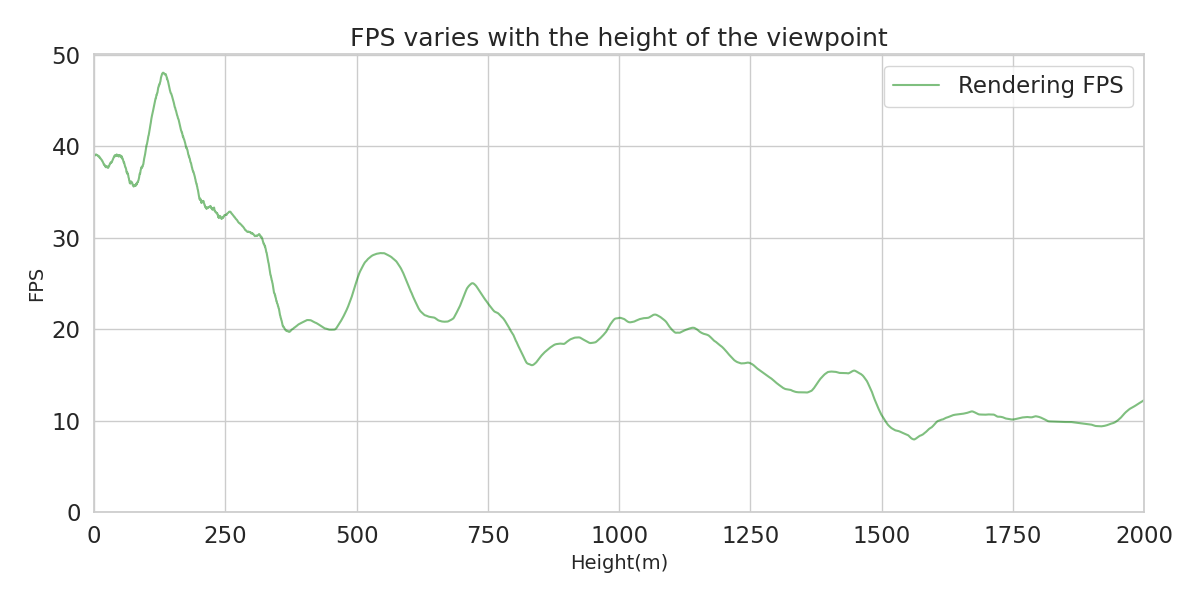}
    \caption{FPS of the basic rending pipeline. As the viewpoint shifts from the ground to a higher altitude and farther away, the height increases, the scene becomes larger, and the FPS decreases accordingly.}
    \label{fig:FPS_from_near_to_far}
\end{figure}

The current 3DGS rendering pipeline, which renders images for two eyes alternatively, is insufficient to support the frame rate required by the immersive VR experience (The minimum requirement of 72 FPS per eye means at least 144 FPS in total. In the following text, FPS refers to the frame rate for both eyes). We conducted experiments on a structured 3DGS model OctreeGS \cite{ren2024octree}, tracing the FPS and analyzing the inference time overhead, as shown in Figure~\ref{fig:FPS_from_near_to_far}. FPS drops as the perspective is elevated. A significant portion of the inference time is spent in the derivation stage, which represents a major overhead. Just as large language models use a KV cache \cite{zheng2024sglang} to accelerate inference, we considered using a cache to speed up 3D model inference. Rendering is usually a continuous process with a lot of overlap between consecutive frames. The overlapping areas reuse data from previous frames, resulting in little color differences. A recent paper \cite{ujjainkar2024exploiting} mentioned that slight color variations are difficult to discern for the human eye. We design an elastic rendering framework with a dynamic cache that stores previous data to accelerate rendering.

In this paper, we propose the GS-Cache framework, a computation framework for large-scale Gaussian splatting rendering, where the scene contains an area that reaches the city-level scale of several square kilometers, to achieve the real-time rendering frame rate requirements for immersive VR experience in binocular 2K resolution head-mounted displays (HMD), as shown in Figure~\ref{fig:system overview}. The middle part of this architecture diagram represents the main structure of the entire framework, the right side shows its elastic parallel scheduler structure, and the left side illustrates the cache-centric rendering pipeline structure. The elastic parallel scheduler schedules GPU resources dynamically, which can steady the FPS and avoid resource waste. For structured 3DGS models, we transform the original pipeline into the cache-centric pipeline, which aims to improve rendering speed based on the principles of de-redundancy and reuse. Additionally, aiming at the bottleneck stages in general computing patterns of the structured Gaussian derivation rendering pipeline, we introduce some dedicated CUDA~\cite{ghorpade2012cuda} kernels for further acceleration, which enhance the frame rate performance of real-time rendering during long-time rendering.

The process of rendering a 3D reconstruction scene involves inference and transformation of the learned 3D spatial features, which makes the conventional computing frameworks that focus on one-dimensional or two-dimensional features such as text and images weak in related tasks, such as Pytorch~\cite{Adam2019Pytorch}, Tensorflow~\cite{Martin2016Tensorflow}, JAX~\cite{Bradbury2018Jax}, etc. These deep learning frameworks are versatile and scalable, enough to implement the basic computing pipeline of neural rendering methods such as NeRF and 3DGS. Still, it is challenging to achieve ease of use for further development in rendering applications, and there is a lack of dedicated operators to support sparse computing in high-dimensional space, resulting in the computing speed of the rendering pipeline being unable to achieve real-time rendering. A series of dedicated frameworks for neural rendering, such as NeRFStudio~\cite{Matthew2023Nerfstudio} and Kaolin-Wisp~\cite{Takikawa2022Kaolin}, have improved the ease of use for model structure experimental research through modularization; and dedicated operator libraries for sparse computing, such as Nerfacc~\cite{Li2023Nerfacc}, have improved the overall rendering speed by accelerating some stages in the NeRF computing pipeline. These works have built a strong community influence and quickly promoted related work on neural rendering such as NeRF and 3DGS, and expanded the applications based on neural rendering. However, the rendering speed still makes it difficult to support the real-time rendering frame rate requirements for immersive VR experiences in large-scale scenes. GS-Cache framework provides a new solution from the perspective of computing systems compatible with various rendering pipelines based on Gaussian derivation strategies. The optimized computing pipeline eliminates the computing redundancy, performs effective computing reuse for immersive VR experience, and flexibly schedules GPU computing resources during the rendering process to ensure stable and high rendering frame rates and optimize the energy efficiency of consumer-grade GPU resources. It accelerates main computing bottlenecks in the pipeline through dedicated CUDA kernels, further improving the performance of VR rendering.

Our main contributions include:

\begin{itemize}
\item A cache-centric computation de-redundancy rendering pipeline that effectively eliminates redundancy in stereo continuous rendering, enabling dynamic cache depth that balances performance and quality.
\item A multi-GPU elastic parallel rendering scheduler that dynamically allocates consumer-grade GPU resources, ensuring stable and high rendering frame rates while enhancing energy efficiency.
\item An end-to-end rendering framework designed for immersive VR experiences, the first holistic system that meets binocular 2K photo-realistic quality rendering requirements of 72 FPS for aerial views and 120 FPS for street views in city-level scenes with dedicated efficient CUDA implementation.
\end{itemize}

\section{Related Work}

Our work focuses primarily on the real-time and photo-realistic rendering of large-scale Gaussian splatting scenes, encompassing city-level scenes with several square kilometers. Although novel view synthesis based on neural rendering has made significant achievements in various applications in recent years, there remains a gap in meeting the demands of rendering performance, quality fidelity, and computational efficiency required for VR rendering. We provide a brief overview of the most relevant works, focusing on real-time photo-realistic rendering, large-scale novel view synthesis, and rendering framework optimizations.

\textbf{Real-Time Photo-realistic Rendering} VR rendering is computationally expensive, requiring high-speed and high-quality real-time rendering, which may be hindered by quality degradation and latency overhead in the general rendering pipeline\cite{song2023nerfplayer}. To achieve high-fidelity rendering with minimal latency under relatively low computation resources, various optimization methods have been proposed. Foveated rendering is a rendering accelerated method, the pioneer work~\cite{guenter2012foveated} providing a foundational theory and approach, the subsequent works~\cite{krajancich2021perceptual,mantiuk2021fovvideovdp,tursun2019luminance} etc., exploring different enhancements and applications. Leveraging eye-tracking technology, foveated rendering tends to allocate more computational resources in rendering the focus area of the images while less the periphery area~\cite{wang2023foveated}. 
To speed up the neural rendering like NeRF, in order to fulfill requirements for real-time rendering, including VR rendering, some works have shifted from pure implicit neural representation towards hybrid or explicit primitive-based neural representations and hardware-based acceleration~\cite{chen2023mobilenerf,tang2023delicate,hedman2021baking}. VR-NeRF~\cite{xu2023vr} achieves high-quality VR rendering using multiple GPUs for parallel computation, RT-NeRF~\cite{li2022rt} realize real-time VR rendering both on cloud and edge devices through efficient pipeline and dedicated hardware accelerator. Re-ReND~\cite{rojas2023re} presents a low resource consumption real-time NeRF rendering method available on resource-constrained devices.~\cite{yu2021plenoctrees, reiser2023merf, yariv2023bakedsdf} distill a pretrained NeRF into a sparse structure, enhancing the real-time rendering performance. 
Different from the aforementioned methods, to speed up neural rendering like 3DGS, another strategy for rendering acceleration involves model pruning and structuring for redundancy removal and effective spatial representation. Methods include ~\cite{fan2023lightGaussian, lee2024compact, lin2024rtgs} pruning Gaussians and reducing model parameters after reconstruction to accelerate the rendering pipeline. Scaffold-GS~\cite{lu2024scaffold} organizes Gaussians using a structured sparse voxel grid and attaches learnable features to each voxel center as an anchor, Octree-GS~\cite{ren2024octree} further employs a structured octree grid for anchors placement. 


\textbf{Large 3D Model Inference} Neural reconstruction and rendering are also attributed to Novel View Synthesis. In large-scale scenes, it has been a long-standing problem in research and engineering. First of all, the fidelity of large-scale rendering is directly contingent upon the quality of the underlying 3D representation models, particularly when reconstructed from real-world scenes. Large-scale scene reconstruction primarily utilizes a divide-and-conquer strategy through scene decomposition methods to expand the capabilities of the model~\cite{tancik2022block, turki2022mega}, while Zip-NeRF~\cite{barron2023zip} and Grid-NeRF~\cite{xu2023grid} better refined the effectiveness and performance of representation for the large-scale scene. Except for the NeRF-based methods ~\cite{park2021instant} extracts semantic information from street-view images and employs panoramic texture mapping method in large-scale scenes novel view synthesis for realism reproduction. To ensure that novel view synthesis for VR real-time rendering maintains a stable frame rate under large-scale scenes, an effective method is the Level of Detail (LoD) strategy.  Guided by heuristic rules or specific resource allocation settings, LoD dynamically adjusts the level of detail layers rendered in real-time ~\cite{Luebke2012LevelOD}. ~\cite{takikawa2021neural} first introduced the concept of LoD into neural radiance fields and neural signed distance, Mip-NeRF~\cite{barron2021mip} and Variable Bitrate Neural Fields~\cite{takikawa2022variable} applying it in the context of multi-scale representation and geometry compressed streaming. LoD has also been employed in Gaussian-based representations, Hierarchy-GS~\cite{kerbl2024hierarchical} designed a hierarchical structure for multi-resolution representation to improve rendering speed. Other large-scale scene reconstruction and rendering works~\cite{xu2023vr, lu2024scaffold, ren2024octree, liu2025citygaussian} have also adopted LoD to accelerate the rendering pipeline.


\textbf{Rendering Framework Optimization} In large-scale novel view synthesis and city-level scene rendering, the stability of high-speed rendering frame rates remains an intractable problem due to variations such as viewpoint and field of view (FOV), as well as the limitations of computational resources. However, little research has focused on optimizations for large-scale VR rendering from the perspective of a computation system, and most existing methods concentrate primarily on mesh-based rendering rather than neural rendering pipelines. MeshReduce ~\cite{jin2024meshreduce} optimizes communication strategy and efficiently converts the scene geometry into the meshes without restraints from computation and memory, yet the stability of rendering frame rates is still difficult to maintain. RT-NeRF~\cite{li2022rt} employs a hybrid sparse encoding method and proposes a NeRF-based storage optimization besides its dedicated hardware system. Post0-VR~\cite{wen2023post0} leverages data similarities to accelerate rendering by eliminating redundant computations and merging common visual effects into the standard rendering pipeline systematically. ~\cite{malkin2020cuda} utilizes shared memory and data reuse to enhance the performance of foveated rendering. 

Our work introduces a novel end-to-end rendering framework for large 3DGS models. Optimizations are applied by an innovative GPU scheduling method, a cache-centric rendering pipeline specifically tailored for Gaussian-based rendering, and dedicated CUDA kernels to stabilize high-speed rendering across immersive VR experiences.

\section{Rendering Pipeline and Framework Design}

GS-Cache is an innovative and holistic rendering framework designed to support the real-time rendering of large-scale 3D scene (3DGS) models at the city level. It enables users to roam in aerial or street views in binocular 2K resolution, achieving an average frame rate exceeding 72 FPS. Given the challenges associated with the real-time photo-realistic rendering of large-scale 3DGS models, particularly on VR applications, we have developed a scheduling framework that supports elastic parallel rendering. Aiming at the Gaussian derivation rendering pipeline patterns, we also propose an efficient cache-centric rendering pipeline with a dynamic cache strategy that maintains rendering quality.

\subsection{Rendering Patterns and Pipeline Bottlenecks}
3DGS represents the structure and color of a scene using a series of anisotropic 3D Gaussians, rendering through rasterization. Structured Gaussian derivation methods use fewer anchors Gaussians and generate more 3D Gaussians from anchors to save GPU resources. 

\textbf{Rendering Patterns} In a point cloud, the position coordinates of each element serve as the mean $\mu$, generating the corresponding 3D Gaussian for differential rasterization rendering:

\begin{equation}
    G( x ) = \exp(-(x-\mu)^T\Sigma^{-1}\frac{(x-\mu)}{2})
\end{equation}

\begin{equation}
    \Sigma = RSS^TR^T
\end{equation}

where $x$ represents any position in the scene space, and $\Sigma$ denotes the covariance matrix of the 3D Gaussian. $\Sigma$ can be decomposed into a rotation matrix $R$  and a scaling matrix $S$ to maintain its positive definiteness. In addition to the mentioned attributes, each 3D Gaussian also includes a color value $c$ and an opacity value $\alpha$. These are used for subsequent opacity blending operations during the rasterization process. While rendering, the 3D Gaussians are first projected onto screen space using the EWA algorithm~\cite{zwicker2002ewa} and transformed into 2D Gaussians, which is a process commonly referred to as $Splatting$. 

In order to make full use of the structured scene priori in the SfM results, some related works have been proposed, such as Scaffold-GS and Octree-GS. Scaffold-GS does not reconstruct directly based on the SfM sparse point cloud but first extracts a sparse voxel grid from the point cloud and constructs anchors at the center of the voxel grids. The anchors contain feature parameters $f$, which are used to derive the neural Gaussian:

\begin{equation}
    \lbrace \mu_j, \Sigma_j, c_j, \alpha_j,\rbrace_{j \in M} = MLP_\mathbf{\theta}\left(f_i,d_{view}\right)_{i \in N}
\end{equation}

\begin{figure}[htb]
 \includegraphics[width=\columnwidth]{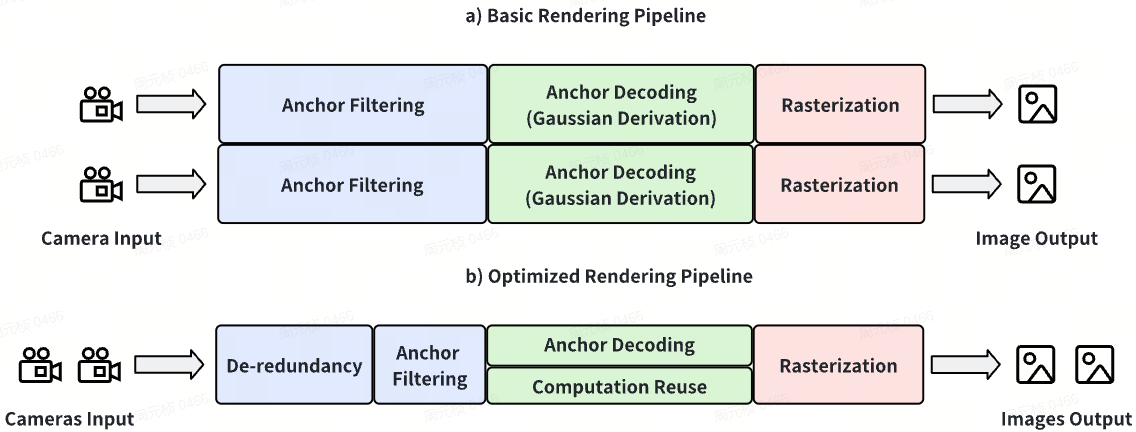}
 \caption{The basic rendering pipeline and our optimized pipeline of structured Gaussian derivation methods}
 \label{fig:rendering pipeline}
\end{figure}

Where \textbf{$\theta$} represents sets of learnable weights of the multi-layer perceptron (MLP), $\mu_j$, $\Sigma_j$, $c_j$, $\alpha_j$ and $s_j$ represent the mean, covariance matrix, color, opacity of the neural Gaussian $j$ derived from anchor $i$ under view direction $d_{view}$. The neural Gaussians will then be used for rasterization, which is no different from native 3D Gaussians. At the same time, the structured placement of the anchors also allows the derived neural Gaussians to be guided by the scene prior, which reduces the redundancy of model parameters and improves robustness in novel view synthesis. Octree-GS goes a step further in the structuring strategy, using an octree to replace the sparse voxel grid to retain multi-resolution and structured scene priori. The multi-resolution grid in the octree makes it possible to construct layers of detail (LoD) in training and then reduce the rendering overhead by setting different detail levels according to distance, expanding the scene scale applicability of the structured Gaussian derivation method. The basic rendering pipeline of Gaussian derivation methods is shown in Figure~\ref{fig:rendering pipeline}.

The structured Gaussian derivation method has higher rendering efficiency than the original randomly distributed Gaussian splatting method; it is still difficult to achieve high-speed rendering (above 72FPS) and ultra-high-speed rendering (above 120FPS) in large-scale scenes such as city-scale scenes. There are two bottleneck stages in the rendering pipeline of structured Gaussian derivation methods:

\begin{itemize}
    \item Gaussian Derivation Stage: Decode the feature parameters of the anchors into neural Gaussian parameters
    \item Gaussian Rasterization Stage: Splat 3D neural Gaussian to 2D and Rasterize neural Gaussians into image 
\end{itemize}

\begin{figure}[htb]
 \includegraphics[width=\columnwidth]{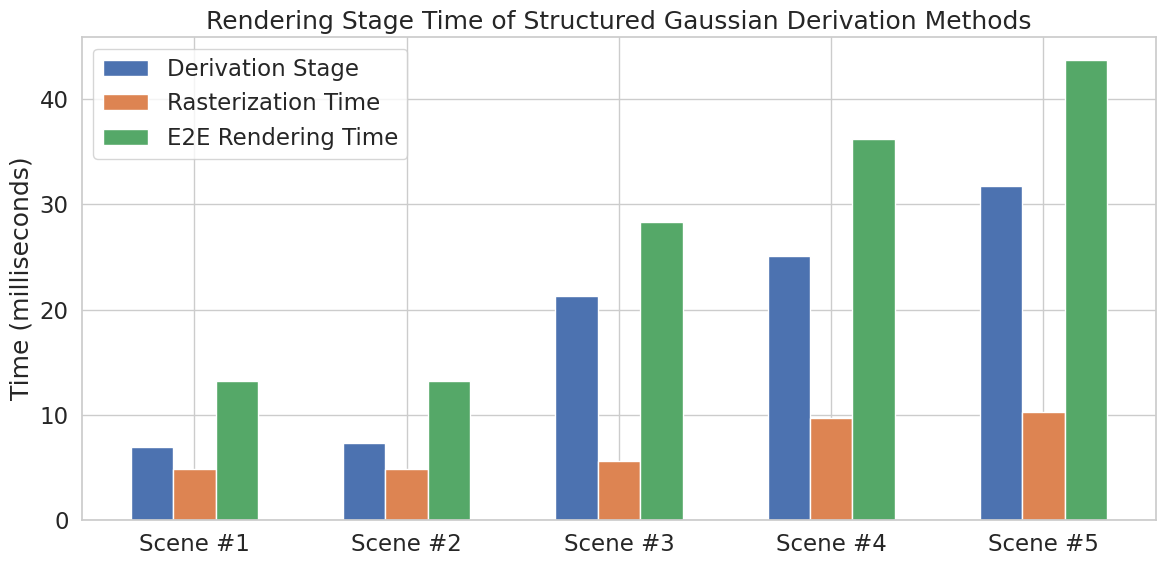}
 \caption{Rendering stage time in different scenes. Although the end-to-end rendering time varies due to the scales of the Gaussian splatting scenes, the time of the derivation stage and the rasterization stage always dominate.}
 \label{fig:computing pattern}
\end{figure}

The derivation stage and rasterization stage are two computationally intensive stages in the rendering pipeline and produce significant temporary GPU memory usage. Therefore, they affect and shape the main computing patterns of the structured Gaussian derivation method and its rendering pipeline in Gaussian splatting scenes of different scales. The following preliminary empirical experiments are shown in the Figure~\ref{fig:computing pattern}. As the scale of the scene increases and the number of anchor feature parameters increases, it further hinders the achievement of the rendering speed required for immersive binocular stereo.

\begin{figure}
    \centering
    \includegraphics[width=1\linewidth]{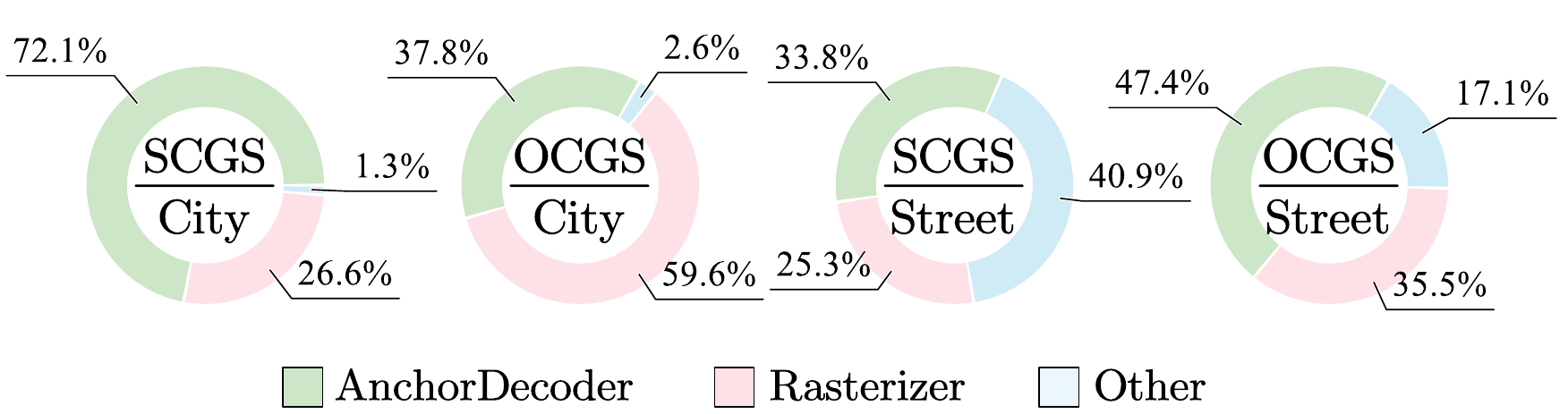}
    \caption{GS model inferring time breakdown. Aside from the Rasterizer operator, other components, including the AnchorDecoder, consume a significant amount of time in the rendering process.}
    \label{fig:gs_model_infer_time}
\end{figure}

We test the proportion of various operators in the model across two common scenarios, as shown in Figure~\ref{fig:gs_model_infer_time}. In the figure, the Rasterizer is the final operator in the rendering process, the AnchorDecoder mainly consists of an MLP, and "Other" refers to some calculations that occur before the AnchorDecoder.

\subsection{Overview of GS-Cache Framework Architecture}
\begin{figure*}[htb]
  \includegraphics[width=\textwidth]{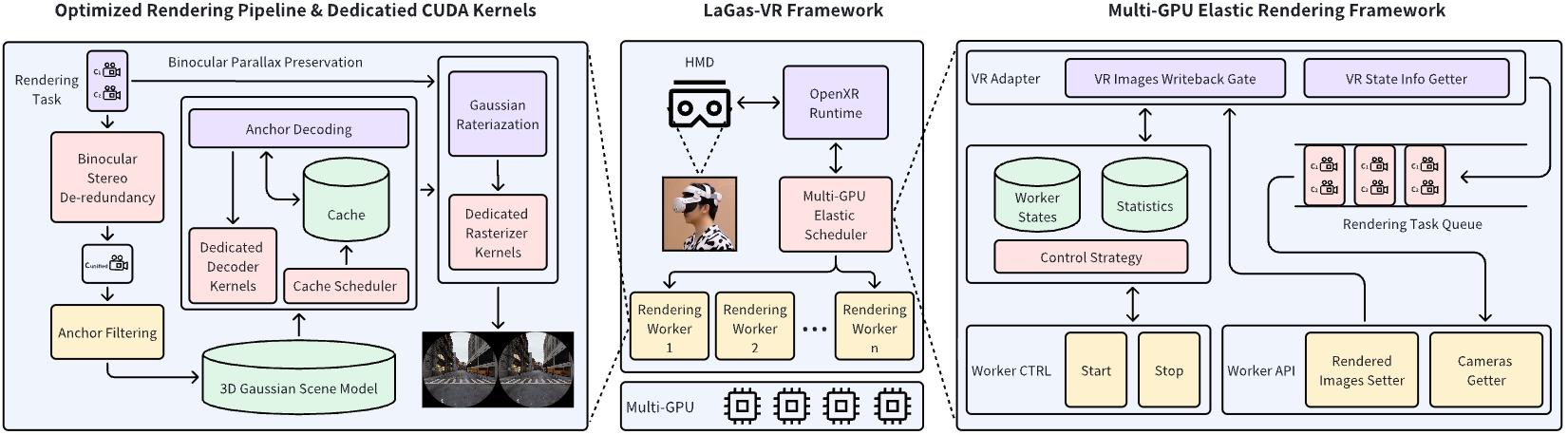}
  \caption{Overview of GS-Cache framework architecture.}
  \label{fig:system overview}
\end{figure*}

The high fidelity of structured Gaussian derivation methods supports the realism of large-scale scene rendering in immersive VR experience, but its basic rendering pipeline is not enough to support the frame rate in large-scale scenes. Therefore, we propose the GS-Cache framework, which realizes the performance requirements of structured Gaussian derivation methods in large-scale scenes from the perspective of computing framework and hierarchical optimization. The GS-Cache framework includes two primary hierarchical structures: a scheduling framework that supports elastic parallel rendering and a de-redundancy rendering pipeline with a cache-centric strategy. As illustrated in Figure~\ref{fig:system overview}.

The scheduling framework is responsible for driving the entire system, facilitating communication with VR applications, and managing the scheduling of rendering workers. It manages worker states and performance statistics, such as the current displaying frame rate and the camera input queue from VR applications. The scheduling framework communicates with the VR applications through a VR adapter, which, alongside the control strategy, enables the initiation or termination of rendering workers while providing interfaces for data transfer.

The de-redundancy rendering pipeline features a multi-level de-redundancy process. Once input data enters the rendering pipeline, it first undergoes a binocular de-redundancy stage, followed by an anchor filtering stage. The filtered anchors are indexed by the model parameters and Gaussian features required for rendering. The cache scheduler then performs the scheduling based on the filtered anchors, loading the necessary model parameters for the subsequent decoding stage. The results of previous decoding computations are updated in the cache, from which the rasterization stage retrieves data for rendering. During continuous rendering, anchors and Gaussian features that are insensitive to viewpoint changes will not require re-decoding. Instead, their decoding results can be directly accessed from the cache.

\section{Cache-centric Computation Reuse Strategy in Rendering}
The core of this framework lies in the cache-centric de-redundancy rendering pipeline, which significantly enhances rendering efficiency. The entire rendering pipeline is built around the cache, which divides it into two main parts. The anchor decoding part generates Gaussian parameters, in which slight changes in perspective only result in little color differences. The cache can optimize and eliminate most of the computations in the anchor decoding part. The rasterizer part involves Gaussian rasterization, as even slight changes in perspective can significantly affect visual perception due to changes in object occlusion. Therefore, this part cannot be accelerated using the cache in order to maintain visual quality.

\begin{figure}
    \centering
    \includegraphics[width=1\linewidth]{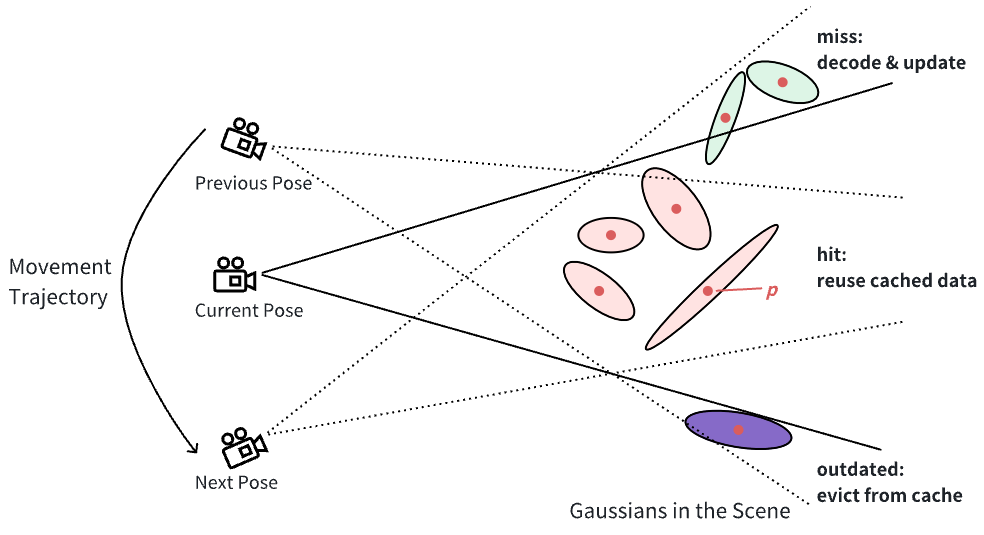}
    \caption{Continuous rendering cache: In continuous rendering, as the viewpoint changes continuously, there is a significant overlap in the view frustum, which can be cached to save computational resources and accelerate rendering.}
    \label{fig:continuous_rendering_cache}
\end{figure}

\textbf{Cache Design} The frame stream generated in real-time rendering is obtained through continuous rendering computations. The input sequence of cameras has spatial proximity, which is also reflected in the fact that the rendering objectives between consecutive frames often overlap. The overlap between consecutive frames in real-time rendering will also cause computation redundancy. Figure \ref{fig:continuous_rendering_cache} illustrates this situation, where the middle section can be cached to accelerate rendering when rendering meets a cache hit. Outdated data is chosen by the cache scheduler and will be evicted. For the continuous rendering of the structured Gaussian derivation method, the computation redundancy is manifested in decoding the same anchor features between consecutive frames, reducing the overall rendering computation efficiency and restraining the upper limit of the frame rate during real-time rendering.


\begin{figure}[htb]
 \includegraphics[width=\columnwidth]{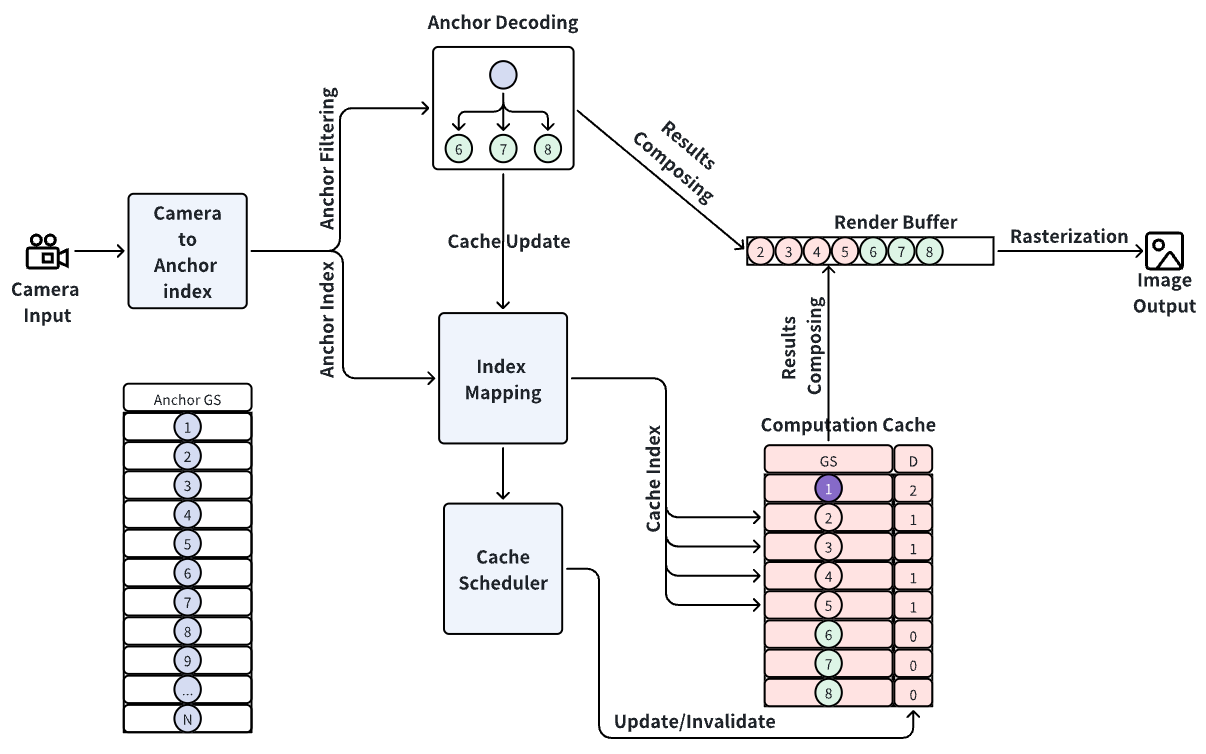}
 \caption{Cache-centric Rendering Pipeline: After the model performs Gaussian parameter decoding, it will be added to the cache. When rendering subsequent frames, if the cache hits, the parameters from the cache will be used.}
 \label{fig:computation reuse rendering pipeline}
\end{figure}

We propose a method based on computation cache to share and reuse intermediate results in the pipeline between multiple frames in the rendering process so that single-frame rendering no longer needs to decode all anchor features but reuse derived Gaussian parameters from the cache and enter the subsequent rasterization stage. The rendering pipeline optimized by computation cache and reuse is illustrated in Figure~\ref{fig:computation reuse rendering pipeline}. The camera input is initially used to compute anchor indices, which are subsequently utilized to filter anchors to be decoded and locate anchors in the cache. The anchor indices are translated to the cache indices by index mapping. The anchor features in current rendering are no longer obtained by complete decoding computations. Still, the results preserved from the previous frames are reused, and the Gaussian parameters not included in the cache are decoded simultaneously. This method removes inter-frame redundancy in the derivation stage for real-time rendering. With an increase in cache depth and a corresponding rise in the reuse rate, both the overall rendering performance and the theoretical upper limit of frames per second (FPS) will undergo further enhancement. This improvement is essential for ensuring the feasibility and stability of high frame rate real-time rendering, which is critical for delivering an immersive experience.

\textbf{Optimizing the Quality–Speed Trade-off} Since the neural Gaussian appearance derived from the anchor point features has a certain viewpoint correlation, the inter-frame reuse of the results of the derivation stage should be restrained by certain conditions to ensure that the correctness of the rendering results within an acceptable range. We design a dynamic cache depth scheduling Algorithm \ref{alg:cache_depth_scheduling} and propose a heuristic cache reuse depth scheduling strategy that adjusts the depth of reuse and cached parameters in subsequent frame renderings according to the intensity of decoding calculations in the derivation stage performed in the current frame rendering. Such a scheduling strategy is not affected by the specific model and pipeline structure. While making full use of the computation cache to accelerate the derivation stage, it can maintain the image rendering quality without a significant decrease or fluctuation due to the use of cached results. Assuming the guiding function of the heuristic strategy is $H$, the expected reuse $depth$ has the following relationship with the anchor set $X$ for decoding computation in the current frame rendering, $X'_k$ means the newly decoded anchors from rendering the previous k frame. This ratio is the update rate:

\begin{equation}
    depth = H(\frac{|X'_k|}{|X|})
\end{equation}

In addition, to avoid the same Gaussian appearance parameters being maintained in the cache for too long and causing significant rendering errors (such as rendering on the trajectories around a centered target), the cache needs to be flushed in time, then cache re-filling and inter-frame reusing need to be performed again. For static cache reuse depth, flushing the cache will cause a rapid change in scene appearance. Therefore, another benefit of introducing dynamic cache depth scheduling is to suppress the appearance of rapid change problems caused by sudden cache flushing.

\begin{algorithm}
\caption{Dynamic Cache Depth Scheduling Algorithm}\label{alg:cache_depth_scheduling}
\begin{algorithmic}[1]
\State Initialize rendering pipeline, set cache depth and guiding function
\While{Receiving camera input}
    \State Anchors indexing and filtering
    \If{Reach max cache reuse depth}
        \State{Invalidate those computation cache line}
    \EndIf
    \If{Anchor duplicate rate == 0\% or first frame}
        \State Decode total anchors into 3D Gaussians
    \ElsIf{Anchor duplicate rate == 100\%}
        \State Reuse total 3D Gaussians in cache
    \Else
        \State Decode new anchors
        \State Update computation cache
        \State Compose new Gaussians and cached Gaussians
    \EndIf

    \State  Configure cache depth based on guiding function and duplicate rate

    \State Update render buffer 

    \State Rasterize render buffer into an image
\EndWhile
\end{algorithmic}
\end{algorithm}

\textbf{Binocular Stereo De-redundancy} Unlike monocular rendering, stereo rendering uses two cameras with sequential alternating method rendering. The overlap of the observation fields further leads to redundancy in stereo rendering; that is, two cameras must render the same objectives of the 3D scene as images under different perspectives. The mere position variance and the large field of view result in significant redundancy in binocular stereo rendering. That means twice anchor decoding and cache visiting.

\begin{figure}[htb]
 \includegraphics[width=\columnwidth]{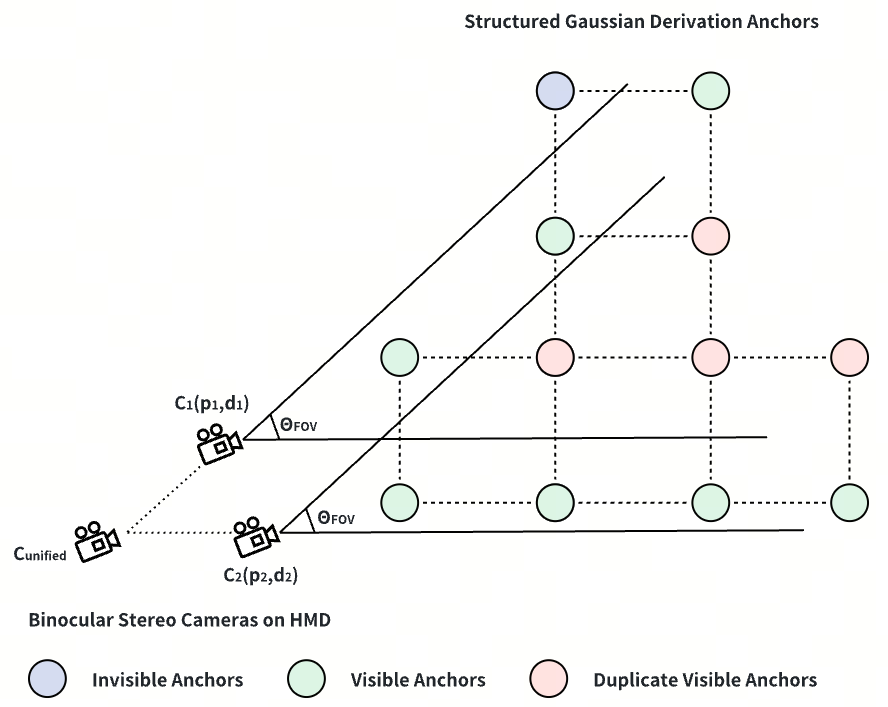}
 \caption{Binocular stereo de-redundancy through uniformed camera in structured Gaussian derivation method.}
 \label{fig:binocular stereo de-redundancy}
\end{figure}


We propose a stereo rendering de-redundancy method suitable for structured Gaussian derivation methods, aiming to eliminate computation redundancy in the derivation stage. The core is to utilize the overlap of the stereo cameras to merge the computation process in the derivation stage so that two cameras can share the Gaussian parameters decoded by a set of anchor features for the subsequent rasterization stage. For binocular stereo rendering, assume that there is a camera group $\mathbf{C}=\lbrace c_1, c_2 \rbrace$, whose positions in the world coordinate system and local Z-axis directions are $\mathbf{P}=\lbrace p_1, p_2 \rbrace$ and $\mathbf{D}=\lbrace d_1, d_2 \rbrace$ , respectively, sharing the same camera intrinsic parameters and expressed as $\mathbf{\theta}_{FOV}$, then the following method can be used to obtain the merged parameters that cover the binocular camera field of view at the same time:

\begin{equation}
    d_{unified} = \frac{\mu_\mathbf{D}}{||\mu_\mathbf{D}||_2}
\end{equation}
\begin{equation}
    p_{unified} = \mu_\mathbf{P} - d_{unified} * \frac{||p_1-p_2||_2}{2*tan(\frac{\mathbf{\theta}_{FOV}}{2})}
\end{equation}

Where $d_{unified}$ and $p_{unified}$ represent the direction and position of the camera, which is equivalent to the combined binocular field of view, and its intrinsic parameters are still $\mathbf{\theta}_{fov}$, as shown in Figure~\ref{fig:binocular stereo de-redundancy}. Therefore, the binocular cameras can simultaneously enter the derivation stage through the equivalent camera and share the results, eliminating the computation redundancy of the derivation stage in the sequential alternating method. Then, the binocular cameras can rasterize the shared Gaussian of the derivation stage independently or in batching to maintain the binocular stereo parallax. It is worth noting that, unlike the double-wide rendering method~\cite{winklehner2007doublewide} in the traditional rasterization of the mesh models, our method merges the multi-channel end-to-end pipeline into one in stereo rendering. This eliminates the redundancy of the sequential alternating method and reduces the number of calls between multiple camera renderings in the derivation stage, thereby improving the performance and frame rate upper limit of stereo rendering.

\section{Multi-GPU Elastic Parallel Rendering Schedule}
In city scenes, since the scenes are large and the Gaussian distribution between the different parts is relatively significant, the performance (FPS) will fluctuate greatly when roaming across the scenes. For example, the FPS will fluctuate greatly from an area with dense high-rise buildings to an open square or from a ground-level view to a high-altitude bird's-eye view. Even though we use cache to reduce the computational load significantly, the rendering FPS decreases as the scene size increases. Therefore, based on the cache-centric rendering pipeline, we further employ elastic parallelism techniques to stabilize the rendering FPS above a predetermined value. That requires that the computing resources be dynamically scheduled according to the changes in the scene. We design an elastic parallel scheduling strategy to alleviate the drastic changes in FPS caused by view changes and achieve stable rendering.

\begin{figure}[htb]
    \centering
    \includegraphics[width=\columnwidth]{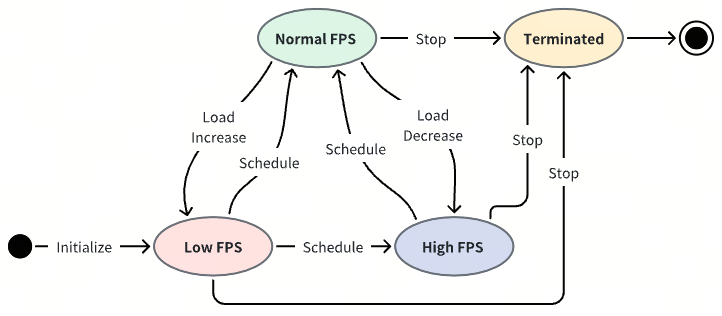}
    \caption{Worker control state transition responding to FPS}
    \label{fig:fps-states}
\end{figure}

We have designed an asynchronous pipeline for VR rendering, as detailed in Algorithm \ref{alg:elastic_parallel}. Rather than directly use the binocular cameras of the VR HMD device for rendering, we put it into a shared queue. In a fixed sampling interval, if the VR HMD device pose changes by more than the threshold, we put the current camera into the shared queue and stamp it with time. The camera data that exceeds a timeout will be discarded by the shared queue. The rendering worker process accesses the shared queue when it finishes rendering the previous camera, takes camera data from the head of the queue, and executes the rendering task. A scheduler is introduced to flexibly schedule the rendering worker process according to the change of FPS. A scheduling strategy is used to achieve a stable rendering. In the strategy, we present a frame rate range [Min-FPS, Max-FPS]. When the FPS is lower than the preset Min-FPS, the scheduler starts a new rendering worker process; when the FPS is higher than $(1+\frac{1}{N_{workers}})$ times of the Max-FPS, a chosen rendering worker process is stopped. As shown in Figure \ref{fig:fps-states}.

Since our elastic parallel rendering is an asynchronous rendering pipeline, inconsistent rendering order may occur due to inconsistent GPU performance. To solve this problem, we synchronize when writing to display. By recording the timestamp of the last frame written to display and comparing the current frame's timestamp, we decide whether the current frame should be written to the GPU memory for display. A simple principle is that frames with earlier timestamps should be displayed first, and we should display rendered frames as quickly as possible. So, under this principle, expired frames will be discarded.

\begin{figure}[htb]
 \centering
 \includegraphics[width=\columnwidth]{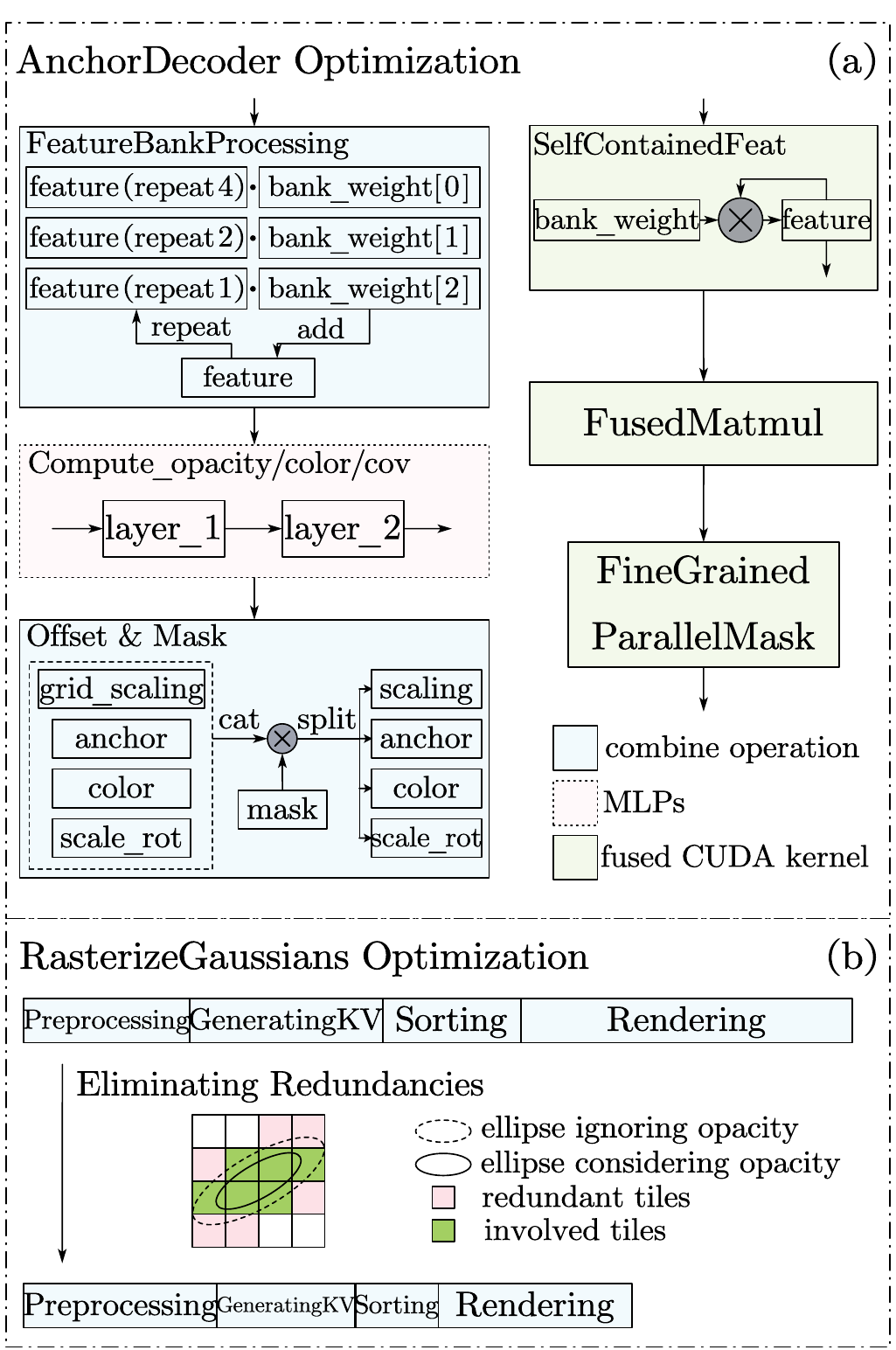}
 \caption{An overview of dedicated CUDA kernels }
 \label{fig:kernel_overview}
\end{figure}

\section{Implementation of model Components}

In structured Gaussian derivation methods, including Scaffold-GS and Octree-GS, the main bottlenecks are the derivation stage (performed by AnchorDecoder) and rasterization stage (performed by RasterizeGaussians), as illustrated by the SCGS and OCGS bars in Figure~\ref{fig:kernel_duration}. Designing dedicated CUDA kernels for these stages significantly improves rendering speed while maintaining a high standard of image output consistency.

\textbf{AnchorDecoder Optimization} The main computational overhead in this stage comes from the combined operators, using 43\% of the stage duration, and the MLPs, using 25\%, as shown in Figure ~\ref{fig:kernel_overview} (a). For the first combine operator, we optimize element-wise multiplication within the kernel, eliminating the need for hard copies of features. For the second combine operator, we replace the merge-and-split operations with a fine-grained parallel method that processes each tensor individually. These optimizations reduce both memory usage and processing time. For the MLPs, our optimization method fuses two layers into a single fused Matmul. For the entire process, precomputing mask indices in the mask computation step can reduce redundant calculations.

\textbf{RasterizeGaussians Optimization}  In addition to general optimizations such as merging memory accesses and precomputation, a significant amount of computational redundancies in the rasterizing pipeline are eliminated through two methods, as shown in Figure ~\ref{fig:kernel_overview} (b). First, when defining the Gaussian distribution, considering the opacity can scale down the size of the ellipse, reducing the area of the Axis-Aligned Bounding Box (AABB) and the number of key-value pairs, thereby reducing the overall computational load. Second, optimizing the AABB's tile coverage determination can eliminate computations of tiles that are completely outside the ellipse's coverage area~\cite{feng2024flashgs}. The redundancy reduction slightly increases the preprocessing duration but significantly reduces the duration of subsequent computational steps.

\begin{algorithm}
\caption{Elastic Parallel Rendering Scheduling Algorithm}\label{alg:elastic_parallel}
\begin{algorithmic}[1]
\State Initialize shared queue, set target FPS and timeout
\While{VR application is tuning}
    \State Obtain current VR device pose information
    \If{HMD device pose change exceeds threshold}
        \State Add camera and timestamp into the shared queue
    \EndIf

    \State Calculate current FPS
    \If{FPS is below target FPS}
        \State Start a new rendering process
    \ElsIf{FPS is above target FPS $(1+\frac{1}{N_{workers}})$}
        \State Choose one rendering process and stop it
    \EndIf

    \For{Each rendered frame}
        \If{Timestamp < Last\_written\_timestamp}
            \State Discard the frame
        \Else
            \State Write the frame to GPU Texture memory
            \State Update Last\_written\_timestamp
        \EndIf
    \EndFor
\EndWhile
\end{algorithmic}
\end{algorithm}

\section{Experiments}
Aiming at an immersive VR experience, we choose the Meta Quest 3 head-mounted display (HMD) as the human-computer interaction interface, which supports a display capability of up to 120FPS with a binocular 2K resolution. The locomotion of the helmet is transmitted through the OpenXR Runtime API, and the rendering results from the other device are streamed to the rendering buffer of the helmet. We use consumer-grade components to build our platform that performs actual rendering, including an Intel i9-14900 CPU, 128GB RAM, and two Nvidia RTX 4090 GPUs connected and communicated through the PCIe 4.0x8 slots. The custom CUDA renderer performs the rendering tasks in the complete framework. The computation results will first be placed in the VRAM on GPUs and finally gathered and streamed to the rendering buffer of the helmet through the OpenXR Runtime API. The GS-Cache rendering framework is deployed on the consumer-grade workstation mentioned above and implemented with PyTorch. All pipelines can use multi-GPU resources to improve the total throughput through the elastic parallel rendering scheduling interface and enable or disable scheduling one of the GPU resources according to the change in rendering frame rate. 

To verify the rendering capabilities of different methods in large-scale scenes, we use the Matrixcity~\cite{li2023matrixcity} dataset to build the target scene used in the experiment. When testing the rendering from the aerial views, the entire 2.7 square kilometers of the city scene is used, while when testing the rendering of the street views, only part of the street scene of the city is used, covering a street range of about 220 meters in length. The sample configurations delivered by Scaffold-GS and Octree-GS on the Matrixcity dataset are trained to 40,000 iterations and obtain scene models with high reconstruction quality, as shown in Table~\ref{tab:scenes}.

\begin{table}[htb]
  \caption{Large-scale Scenes reconstructed from the Matrixcity dataset}
  \label{tab:scenes}
  \scriptsize
  \begin{tabularx}{\linewidth}{l X X X X X}
  \toprule
  \textbf{Scene} &\textbf{View} &\textbf{Scale} &\textbf{PSNR} &\textbf{Images} &\textbf{Anchors} \\
  \midrule
  City(Scaffold-GS) &Aerial &2.7$km^2$ &28.84 &5621 &18,554k \\
  City(Octree-GS) &Aerial &2.7$km^2$ &28.08 &5621 &17,122k \\
  \midrule
  Street(Scaffold-GS) &Street &0.026$km^2$ &30.01 &330 &1,311k \\
  Street(Octree-GS) &Street &0.026$km^2$ &31.20 &330 &2,848k \\
  \bottomrule
  \end{tabularx}
\end{table}

\subsection{Performance Evaluation}
To simulate the experience of urban roaming, we set an aerial ascending curved trajectory in the city scene and a straight street trajectory along the road in the street scene. We take the consecutive keyframes in the trajectory as input for our rendering framework. Each frame corresponds to the head-mounted display's locomotion input, which contains two poses of the binocular stereo cameras on it.

Experiments are conducted on Scaffold-GS and Octree-GS, the SOTA models for modeling cities. The max reuse depth is set to 10 for caching computation results from the previous 10 frames. A series of experiments show that setting it to 10 balances both performance and quality. To compare their optimal performance, both rendering pipelines are tested under single GPU and multi-GPU resources. Besides, we compare our results with CityGS \cite{liu2025citygaussian} and some VR rendering works.

We collect the average FPS for performance evaluation of the rendering pipeline under continuous computing conditions. For a more comprehensive result, we also collect the 99$\%$ percentile FPS. At the same time, the time consumption of the anchor decoding stage and the Gaussian rasterization stage in the rendering pipeline, where computation optimization and dedicated CUDA kernels mainly take effect, are collected. The results are shown in Table~\ref{tab:city performance} and Table~\ref{tab:street performance}.

\begin{table*}[htb]
  \centering
  \caption{Rendering performance comparison on Matrixcity city scene}
  \label{tab:city performance}
  \scriptsize
  \begin{tabular}{lccccccc}
  \toprule
  \textbf{Methods} &\textbf{AVG. FPS} &\textbf{99$\%$ FPS} &\textbf{Decoding(ms)} &\textbf{Rasterization(ms)} &\textbf{AVG. Speedup Gain}\\
  \midrule
  Scaffold-GS(Origin) &27.24 &13.33 &28.89 &10.63 &-\\
  Octree-GS(Origin) &18.04 &11.50 &18.49 &14.10 &-\\
  Scaffold-GS(Our) &55.81 &37.28 &11.34 &5.42 &2.05\\
  Octree-GS(Our) &44.55 &25.81 &6.39 &4.35 &2.47\\
  \midrule  
  Scaffold-GS(Origin w/ elastic) &50.78 &29.07 &14.19 &5.19 &1.86\\
  Octree-GS(Origin w/ elastic) &42.38 &30.98 &8.01 &5.92 &2.35\\
  Scaffold-GS(Our w/ elastic) &109.80 &78.29 &5.57 &3.01 &4.03\\
  Octree-GS(Our w/ elastic) &96.46 &73.27 &2.82 &2.29 &5.35\\
  \bottomrule
  \end{tabular}
\end{table*}

\begin{table*}[htb]
  \centering
  \caption{Rendering performance comparison on Matrixcity street scene}
  \label{tab:street performance}
  \scriptsize
  \begin{tabular}{lcccccc}
  \toprule
  \textbf{Methods} &\textbf{AVG. FPS} &\textbf{99$\%$ FPS} &\textbf{Decoding(ms)} &\textbf{Rasterization(ms)} &\textbf{AVG. Speedup Gain}\\
  \midrule
  Scaffold-GS(Origin) &43.91 &8.35 &8.67 &13.38 &-\\
  Octree-GS(Origin) &54.26 &21.19 &8.62 &6.63 &-\\
  Scaffold-GS(Our) &80.24 &14.80 &1.61 &9.88 &1.83\\
  Octree-GS(Our) &113.97 &44.74 &2.25 &4.37 &2.10\\
  \midrule
  Scaffold-GS(Origin w/ elastic) &88.59 &18.24 &4.36 &6.57 &2.02\\
  Octree-GS(Origin w/ elastic) &111.98 &44.38 &4.16 &3.11 &2.06\\
  Scaffold-GS(Our w/ elastic) &148.30 &33.34 &0.93 &5.59 &3.38\\
  Octree-GS(Our w/ elastic) &203.16 &93.69 &1.18 &2.54 &3.74\\
  \bottomrule
  \end{tabular}
\end{table*}

Under a single GPU, compared to the baseline pipeline, the optimized rendering pipeline has an average frame rate performance improvement of 2x. The city scene has a higher speedup gain because the number of anchors involved in decoding and the number of Gaussians involved in rasterization differ in the two scenes. Our framework demonstrates its advantages more clearly in larger scenes. When the number of anchors in the scene is low, the decoding stage will become the main bottleneck in rendering, rather than the rasterization stage, in which the gain will be more significant. The worst performance is relatively close in all pipelines, the real bottleneck in processing the input sequence. However, the stable frame rate represented by 99$\%$ FPS also has a performance improvement of more than 2x, indicating that the performance changes between the optimized pipeline and the baseline pipeline are not instantaneous but rather a steady increase in performance throughout the entire continuous rendering process. Under multi-GPU, the average frame rate of the optimized rendering pipeline can not only achieve the minimum 72FPS requirement for an immersive VR experience. Still, it can even reach more than 120FPS in street scenes, exceeding the maximum refresh rate limit of the head-mounted display. The baseline pipeline, with the support of multi-GPU resources, can also achieve the minimum 72FPS requirement for immersive VR experience in street scenes.

\begin{figure}[htb]
 \includegraphics[width=\columnwidth]{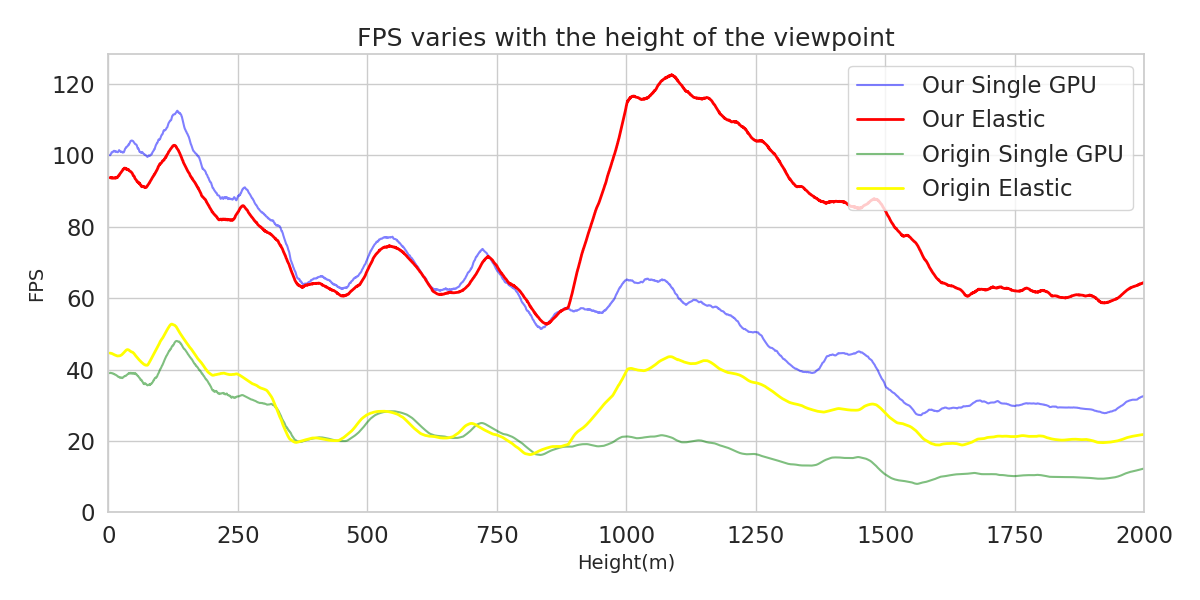}
 \caption{Rendering frame rate on the trajectory w.r.t height. Elastic scheduling can maintain acceptable FPS}
 \label{fig:zoom out frame rate}
\end{figure}

As shown in the Figure~\ref{fig:zoom out frame rate}, to verify the scheduling capability of our multi-GPU elastic rendering method, we set an additional zoom-out trajectory on the city scene, pointing to the center of the scene and moving away from it, and compare the frame rate changes of the Octree-GS optimized rendering pipeline and the baseline pipeline under fixed single GPU resources and elastic multi-GPU resources. When the camera gradually moves away from the center of the scene along the zoom-out trajectory, more anchors will be involved in the decoding stage of the rendering pipeline. Then more Gaussians will be derived and involved in the rasterization stage. In the case of static resource scheduling, the end-to-end throughput of the rendering pipeline will gradually degrade, even below the minimum frame rate requirement for an immersive VR experience. However, an elastic resource scheduling strategy can allocate only part of the GPU resources within the acceptable performance fluctuation range and allocate more GPU resources when the performance drops outside, enabling another GPU to participate in rendering and restoring the performance to the acceptable range. This can improve the energy efficiency of the rendering framework and maximize the utilization of resources scheduled for computing tasks. Still, it is also friendly to target applications on consumer-grade devices. We set the performance acceptable range of the optimized pipeline to a minimum of 60 FPS, the minimum requirement for a smooth VR experience, but cannot achieve an immersive VR experience. At the same time, the performance acceptable range of the baseline pipeline is set to a minimum of 20 FPS, which is the minimum requirement for real-time rendering. Frame rates lower than this will not even meet the real-time application requirements. 

In contrast to methods such as VR-NeRF~\cite{xu2023vr}, RT-NeRF~\cite{lin2024rtgs}, and VR-GS~\cite{jiang2024vrgs}, which have made significant contributions to the VR rendering of 3D neural scenes, these methods have primarily focused on small-scale scenes. The model sizes employed in these methods differ by an order of magnitude from those in our experiments, and their average rendering frame rates are significantly below 72 FPS. Our solution, however, substantially outperforms existing methods in terms of performance while maintaining rendering quality. At the same settings, from the viewpoint of 500 meters in height, the FPS of GS-Cache is double that of CityGaussian \cite{liu2025citygaussian}.

\subsection{Quality Evaluation}

\begin{figure}[htb]
 \includegraphics[width=\columnwidth]{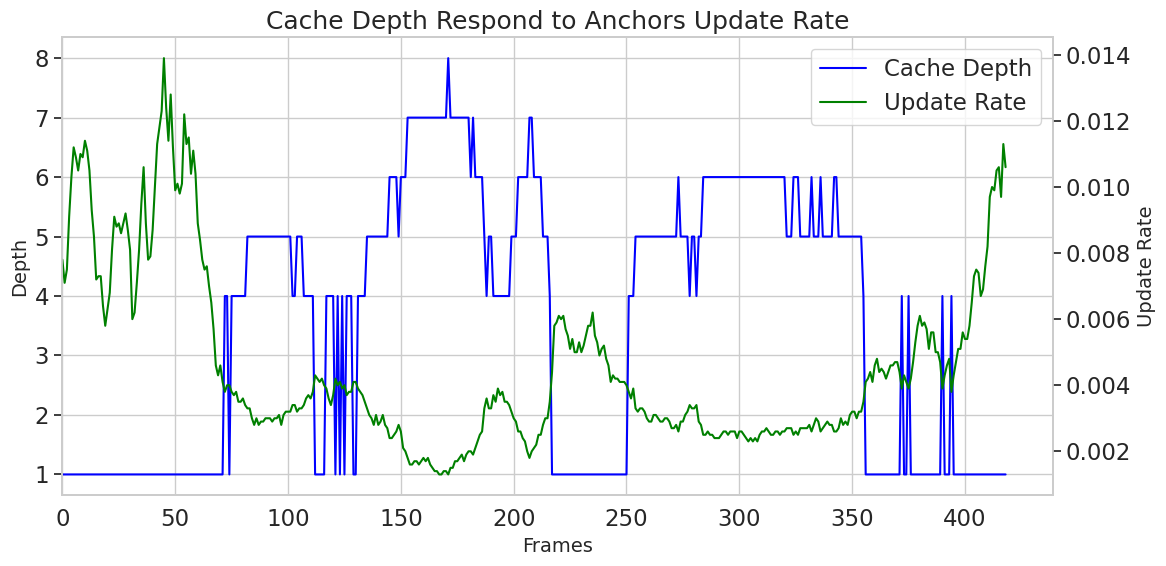}
 \caption{Dynamic cache depth. Due to the scheduling strategy for quality, it may cause rapid shrinkage of depth when the update rate fluctuates, but overall speedup can still be achieved.}
 \label{fig:Dynamic cache depth}
\end{figure}

Performance improvements in rendering pipelines are often accompanied by trade-offs in quality. The rendering pipeline optimization method we proposed includes computation de-redundancy and computation reuse to maintain the pipeline structure unchanged and optimize the process centered on rendering quality. Due to the peculiarities of multi-camera setups in binocular stereo, the impact of redundancy removal on rendering quality is not significant. In the meantime, reuse requires a dynamic cache depth scheduling strategy to control fluctuations in rendering quality. In our experiments, we use a linear guidance function for responding to intensity changes in the decoding stage and schedule reuse depth in subsequent frame renderings. The linear response is matched with the motion features for the constant speed movement of the rendered trajectory in the experiment. Figure~\ref{fig:Dynamic cache depth} illustrates how the cache depth adjusts to maintain rendering quality during the rendering process. The update rate refers to the percentage of anchors that are decoded and updated to the cache and is equivalent to the cache miss rate. As the cache depth increases, the cache miss rate decreases, resulting in a corresponding reduction in the update rate. For the movement with acceleration and with staged speed changes, exponential response and staged response are needed to match the reuse depth and motion features. Our optimization methods do not involve modifications on the rendering pipeline, are transparent to the original rendering process of the Gaussian derivation method, and are compatible with pipelines without LOD (e.g., Scaffol-GS) and pipelines containing LOD (e.g., Octree-GS). 

\begin{table}[htb]
  \caption{Rendering quality comparison on Matrixcity city and street scenes}
  \label{tab:quality loss}
  \scriptsize
  \begin{tabularx}{\linewidth}{l X X X X}
  \toprule
  \textbf{Methods} &\textbf{MSE↓} &\textbf{PSNR↑} &\textbf{SSIM↑} &\textbf{LPIPS↓}\\
  \midrule
  Scaffold-GS(City) &0.00116 &38.36 &0.98 &0.022 \\ 
  Octree-GS(City) &0.00136 &35.38 &0.98 &0.024 \\ 
  \midrule
  Scaffold-GS(Street) &0.00155 &32.68 &0.98 &0.018 \\
  Octree-GS(Street) &0.00038 &35.53 &0.99 &0.012 \\  
  \bottomrule
  \end{tabularx}
\end{table}

We evaluate the quality difference caused by computation optimizations between images rendered by the optimized and baseline pipelines. As shown in the Table~\ref{tab:quality loss}. It is worth noting that the mean square error(MSE) and peak signal-to-noise ratio(PSNR) can only reflect the absolute difference in pixel values between the images but not the relative difference in perception. Therefore, it is also necessary to refer to metrics such as the structural similarity index(SSIM)\cite{wang2004ssim} and the learned perceptual image patch similarity(LPIPS)\cite{zhang2018lpips}. It is generally believed that when PSNR surpasses 30, the visual difference between the two images is difficult to perceive by the human eye. SSIM and LPIPS need to be over 0.9 and no more than 0.1, respectively.

\subsection{Ablation Evaluation}

In addition to evaluating the overall performance and quality changes of the GS-Cache rendering framework, we also conduct ablation experiments on the performance impact of different optimization methods for the rendering pipeline. Based on the rendering trajectories from the aerial and the street, pipelines that ablated different optimization methods are tested for rendering. We compare the end-to-end frame rate and stage time consumption, and record the highest memory usage in the complete rendering process, as Table~\ref{tab:city ablation} And Table~\ref{tab:street ablation} shown. We can also calculate the speedup of different methods to achieve full performance through ablation.

\begin{table*}[htb]
  \centering
  \caption{Rendering ablation comparison on Matrixcity city scene.}
  \label{tab:city ablation}
  \scriptsize
  \begin{tabular}{lcccccc}
  \toprule
  \textbf{Methods} &\textbf{AVG. FPS} &\textbf{99$\%$ FPS} &\textbf{Decoding(ms)} &\textbf{Rasterization(ms)} &\textbf{Memory(GiB)} &\textbf{AVG. Speedup Loss}\\
  \midrule
  Scaffold-GS(Our) &56.61 &38.71 &11.28 &5.39 &8.10 &-\\
  Our w/o de-redundancy &34.82 &26.49 &22.72 &5.36 &8.05 &1.62x\\
  Our w/o reuse &31.26 &20.52 &19.48 &5.27 &7.85 &1.81x\\
  Our w/o kernels &38.46 &21.64 &14.15 &10.77 &10.30 &1.47x\\
  \midrule  
  Octree-GS(Our) &41.15 &20.33 &6.89 &4.68 &8.59 &-\\
  Our w/o de-redundancy &24.86 &16.10 &13.27 &4.37 &8.59 &1.65x\\
  Our w/o reuse &22.48 &17.66 &10.09 &4.54 &8.51 &1.83x\\
  Our w/o kernels &31.51 &21.40 &8.41 &12.57 &9.17 &1.30x\\
  \bottomrule
  \end{tabular}
\end{table*}

\begin{table*}[htb]
  \centering
  \caption{Rendering ablation comparison on Matrixcity street scene.}
  \label{tab:street ablation}
  \scriptsize
  \begin{tabular}{lcccccc}
  \toprule
  \textbf{Methods} &\textbf{AVG. FPS} &\textbf{99$\%$ FPS} &\textbf{Decoding(ms)} &\textbf{Rasterization(ms)} &\textbf{Memory(GiB)} &\textbf{AVG. Speedup Loss}\\
  \midrule
  Scaffold-GS(Our) &79.43 &14.59 &1.68 &9.93 &2.12 &-\\
  Our w/o de-redundancy &72.52 &14.59 &3.20 &9.95 &2.16 &1.09x\\
  Our w/o reuse &77.88 &13.60 &1.84 &10.01 &1.96 &1.02x\\
  Our w/o kernels &58.34 &11.43 &2.90 &13.18 &3.67 &1.36x\\
  \midrule
  Octree-GS(Our) &115.88 &47.93 &2.19 &4.34 &2.07 &-\\
  Our w/o de-redundancy &87.85 &44.87 &4.16 &4.25 &2.12 &1.31x\\
  Our w/o reuse &93.02 &41.30 &2.58 &4.35 &2.03 &1.27x\\
  Our w/o kernels &77.76 &33.45 &4.20 &6.36 &2.94 &1.49x\\
  \bottomrule
  \end{tabular}
\end{table*}

In the city scene, computation de-redundancy and reuse significantly impact the pipeline's overall performance, and the speedup of the average frame rate can reach 1.8x. In the street scene, the impact of dedicated CUDA kernels on the overall performance of the pipeline is more significant. This is because, in the city scene, the time of single-frame rendering is concentrated in the decoding stage. In contrast, in the street scenes, the time is concentrated in the rasterization stage, corresponding to the main effects of computation optimization and dedicated kernels. The impact on the average frame rate is also reflected in the 99$\%$ frame rate, which further proves that our optimization methods are essential to improve the performance of the rendering pipeline, which covers the best and worst cases in rendering performance. Removing redundancy in binocular stereo improves the performance of the decoding stage most significantly in different scenes and pipelines. In contrast, dedicated kernels can most significantly improve the performance of the rasterization stage while reducing memory usage. Due to the introduction of the computation cache for Gaussian derivation, the reuse optimization of the rendering pipeline will cause additional memory usage, not exceeding 3$\%$. Ultimately, our optimization methods jointly support the rendering pipeline to meet the performance requirements of immersive real-time VR rendering and form compensation and balance in memory usage.

\begin{figure}[htb]
 \includegraphics[width=\columnwidth]{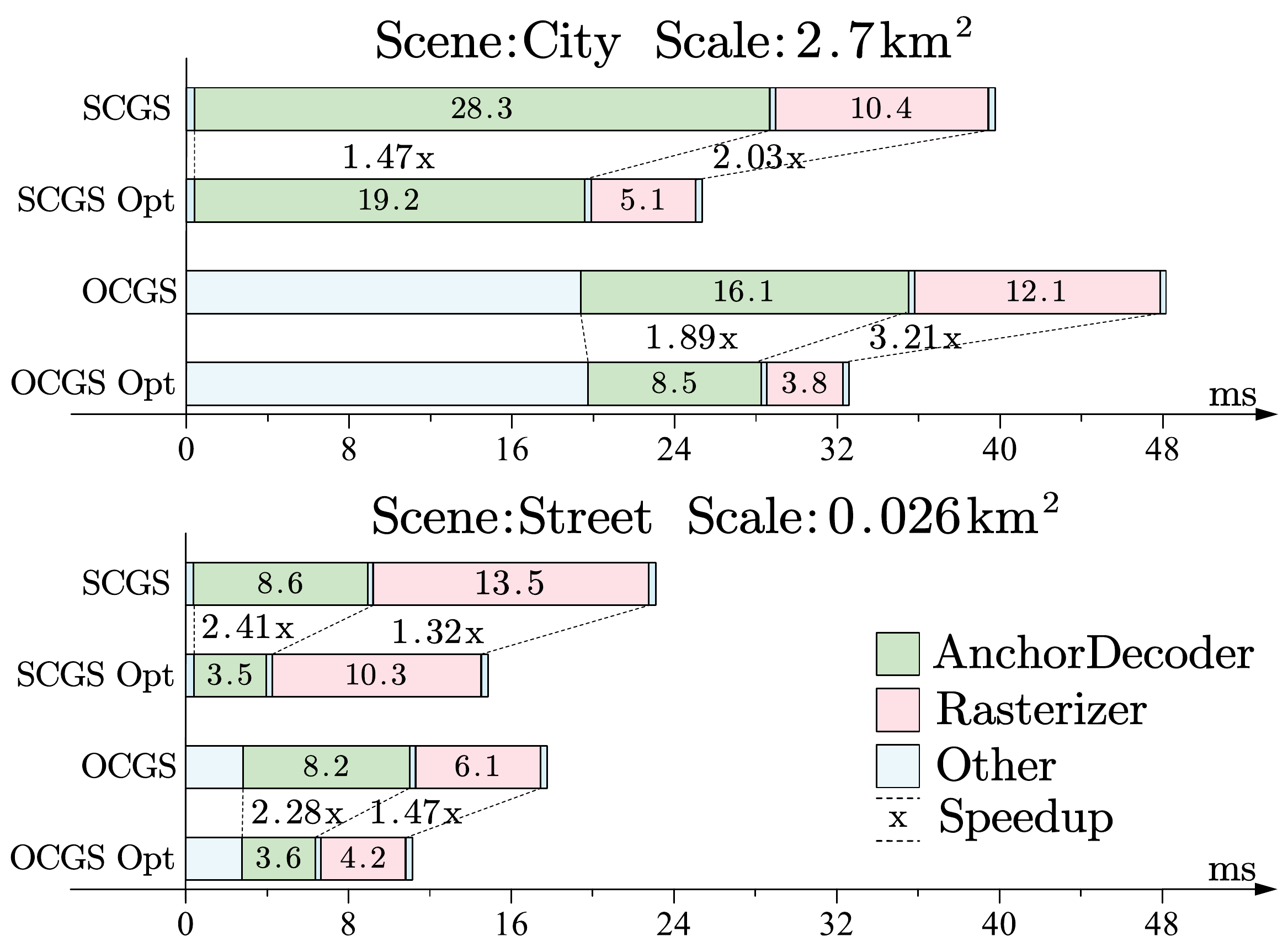}
 \caption{Single-frame rendering kernel duration. Due to the LOD-based model structure and layer-switching strategy applied to the entire scene, Octree-GS introduces significant time before the decoding stage in large-scale scenes.}
 \label{fig:kernel_duration}
\end{figure}

As shown in Figure~\ref{fig:kernel_duration}, replacing only the dedicated CUDA kernel results in a speedup for both AnchorDecoder and Rasterizer. Notably, in the city and street scenes, Rasterizer achieves a higher speedup in the former, while AnchorDecoder achieves a higher speedup in the latter. This is because the city scene spans a broader area and involves a larger scale, with denser and more numerous anchor points and Gaussians. For Rasterizer, the optimization eliminates redundant computations caused by false Gaussian intersections. The optimization yields better results since the denser Gaussians in the city scene result in more false intersections. As for AnchorDecoder, its optimization mainly reduces memory access overhead. In the city scene, the increased density of anchor points raises the computational overhead for AnchorDecoder, making memory access optimizations less effective than in the street scene.
\vspace{-5pt}
\section{Conclusions}
We demonstrate the GS-Cache framework, a rendering framework oriented to structured Gaussian derivation methods, which can achieve real-time rendering of large-scale scenes, including city and street Gaussian reconstruction scenes, meeting the high-speed and high-fidelity requirements of immersive VR experience. We make several key contributions, including the cache-centric de-redundancy rendering pipeline, a rendering framework that supports multi-GPU parallelism and elastic scheduling, and dedicated CUDA kernels for the computational bottleneck stage. In the experiments, we verify that the GS-Cache framework achieves significant performance improvements compared to the baseline methods, and meets the frame rate requirements of binocular 2K resolution of more than 72FPS and more than 120FPS under limited resources such as consumer-grade GPUs, and does not result in significant quality loss.

\bibliographystyle{plain}
\bibliography{main}

\end{document}